\documentclass[journal]{IEEEtran}
\usepackage{graphicx} 
\usepackage{booktabs}
\usepackage[table,xcdraw]{xcolor}
\usepackage{url}
\usepackage{tabularx}
\usepackage{xcolor}
\usepackage{booktabs}
\usepackage{adjustbox}
\usepackage{float}
\usepackage{amsmath}
\usepackage{cite}
\usepackage{subcaption}
\usepackage{graphicx}
\usepackage{algorithm}
\usepackage{algpseudocode}
\usepackage{multirow}

\usepackage{hyperref}

\ifCLASSINFOpdf

\else

\fi

\begin{document}

\title{Enhancing Osteoporosis Detection: An Explainable Multi-Modal Learning Framework with Feature Fusion and Variable Clustering}

\author{Mehdi Hosseini Chagahi, Saeed Mohammadi Dashtaki, Niloufar Delfan, Nadia Mohammadi, Farshid Rostami Pouria, Behzad Moshiri (Senior Member, IEEE), Md. Jalil Piran (Senior Member, IEEE), Oliver Faust

\thanks{\textit{(M. H. Chagahi and S. M. Dashtaki contributed equally to this work.)(Corresponding Authors: B. Moshiri  and M. J. Piran.)}}
\thanks{M. H. Chagahi, S. M. Dashtaki , N. Delfan, and F. R. Pouria are with the School of Electrical and Computer Engineering, College of Engineering, University of Tehran, Tehran, Iran, (e-mail: mhdi.hoseini@ut.ac.ir;
saeedmohammadi.d@ut.ac.ir;
niloufardelfan@gmail.com; farshid.rostami73@ut.ac.ir)}

\thanks{N. Mohammadi is with the Department of Epidemiology, Shiraz University of Medical Science, Shiraz, Iran, (e-mail: nmohammadi.nadia@gmail.com)}

\thanks{B. Moshiri is with the School of Electrical and Computer Engineering, College of Engineering, University of Tehran, Tehran, Iran and the Department of Electrical and Computer Engineering University of Waterloo,
Waterloo, Canada, (e-mail: moshiri@ut.ac.ir)}

\thanks{M. J. Piran is with the Department of Computer Science and Engineering, Sejong University, Seoul 05006, South Korea, (e-mail: piran@sejong.ac.kr)}

\thanks{O. Faust is with the School of Computing and Information Science, Anglia Ruskin University, East Road, Cambridge, UK, (e-mail: oliver.faust@gmail.com)}
}
\maketitle

\begin{abstract}

Osteoporosis is a common condition that increases fracture risk, especially in older adults. Early diagnosis is vital for preventing fractures, reducing treatment costs, and preserving mobility. However, healthcare providers face challenges like limited labeled data and difficulties in processing medical images. This study presents a novel multi-modal learning framework that integrates clinical and imaging data to improve diagnostic accuracy and model interpretability. The model utilizes three pre-trained networks—VGG19, InceptionV3, and ResNet50—to extract deep features from X-ray images. These features are transformed using PCA to reduce dimensionality and focus on the most relevant components. A clustering-based selection process identifies the most representative components, which are then combined with preprocessed clinical data and processed through a fully connected network (FCN) for final classification. A feature importance plot highlights key variables, showing that Medical History, BMI, and Height were the main contributors, emphasizing the significance of patient-specific data. While imaging features were valuable, they had lower importance, indicating that clinical data are crucial for accurate predictions. This framework promotes precise and interpretable predictions, enhancing transparency and building trust in AI-driven diagnoses for clinical integration.

\end{abstract}

\begin{IEEEkeywords}
Osteoporosis detection, Explainable AI, Variable clustering, Transfer learning, Multi-modal learning.
\end{IEEEkeywords}

\IEEEpeerreviewmaketitle

\begin{figure*}[t!]
    \centering
    \captionsetup[subfigure]{labelformat=empty} 
    \begin{subfigure}{0.32 \textwidth}
        \includegraphics[width=0.85\linewidth]{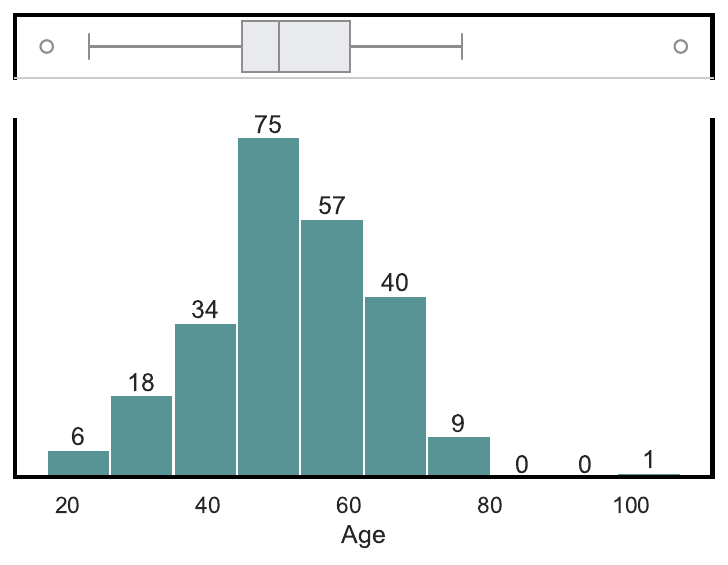}
        \subcaption{} 
    \end{subfigure}
    \begin{subfigure}{0.32 \textwidth}
        \includegraphics[width=0.85\linewidth]{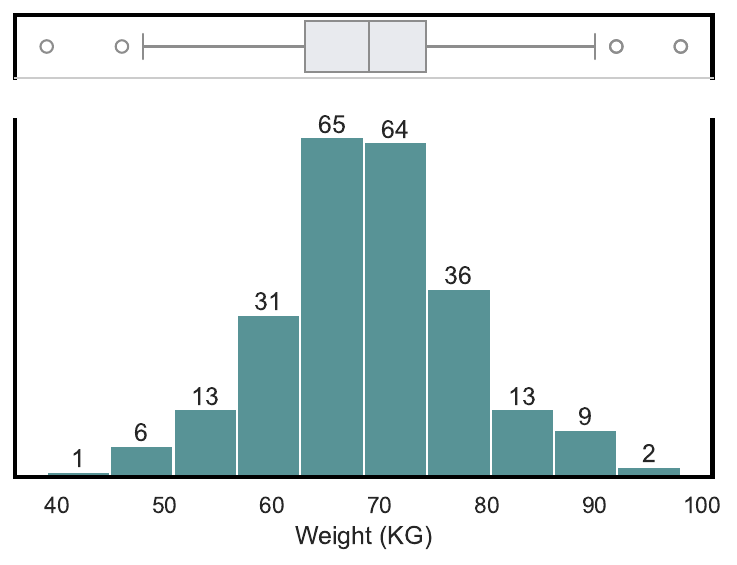}
        \subcaption{} 
    \end{subfigure}
    \begin{subfigure}{0.32 \textwidth}
       \includegraphics[width=0.85\linewidth]{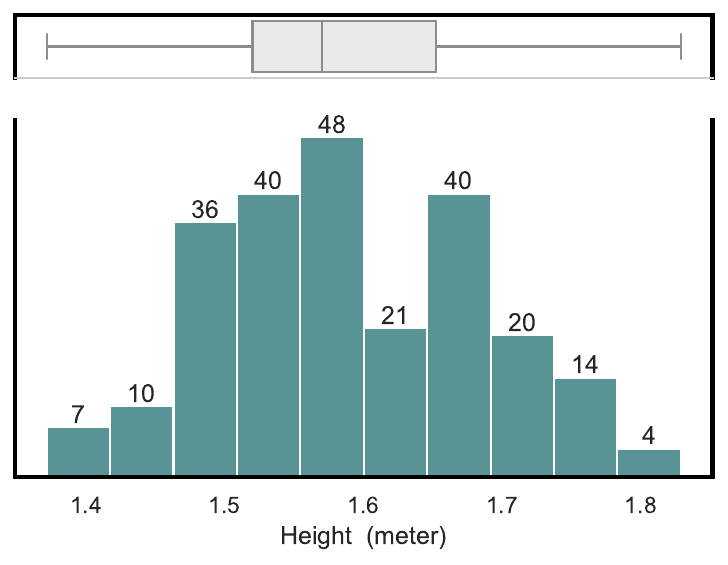}
       \subcaption{} 
    \end{subfigure}
    \begin{subfigure}{0.32 \textwidth}
        \includegraphics[width=0.85\linewidth]{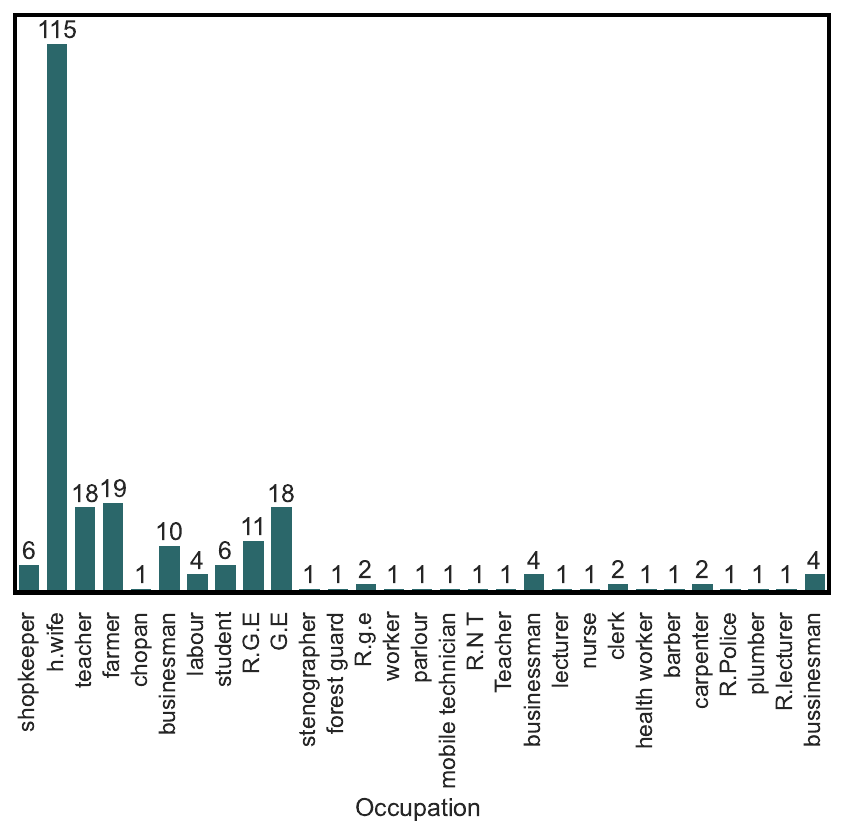}
       \subcaption{} 
    \end{subfigure}
    \begin{subfigure}{0.32 \textwidth}
        \includegraphics[width=0.85\linewidth]{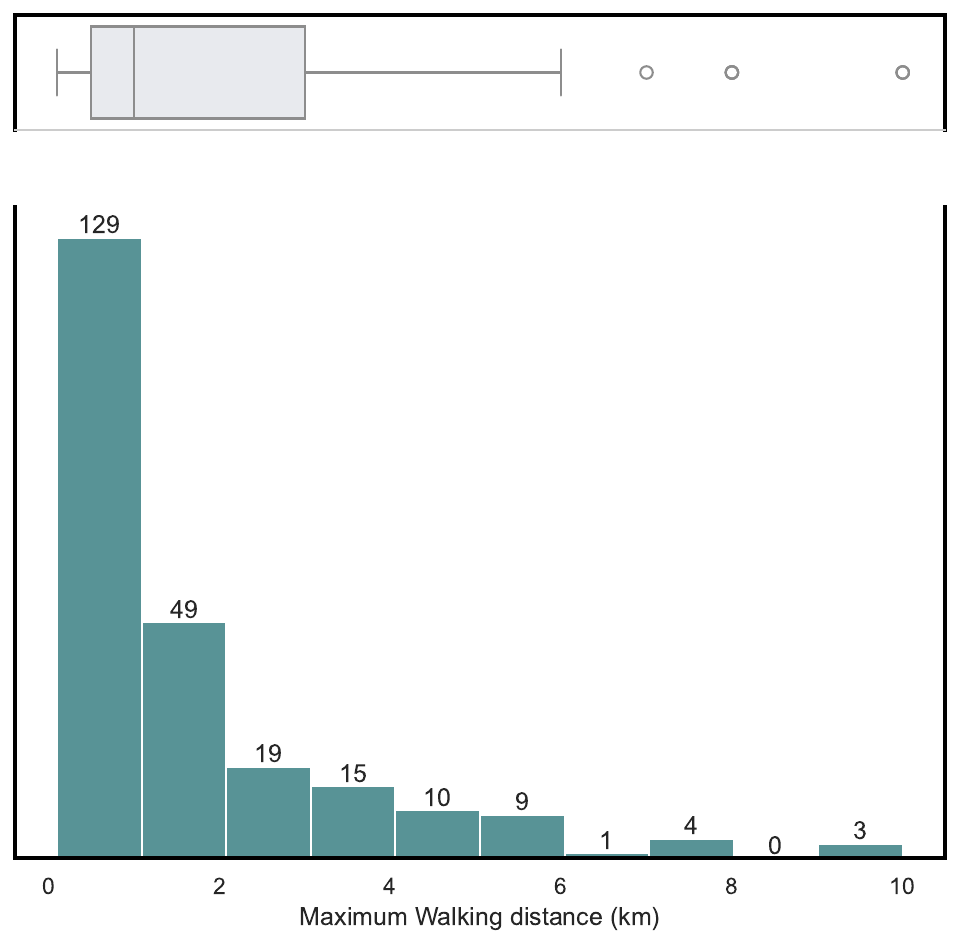}
       \subcaption{} 
    \end{subfigure}
    \begin{subfigure}{0.32 \textwidth}
        \includegraphics[width=0.85\linewidth]{ 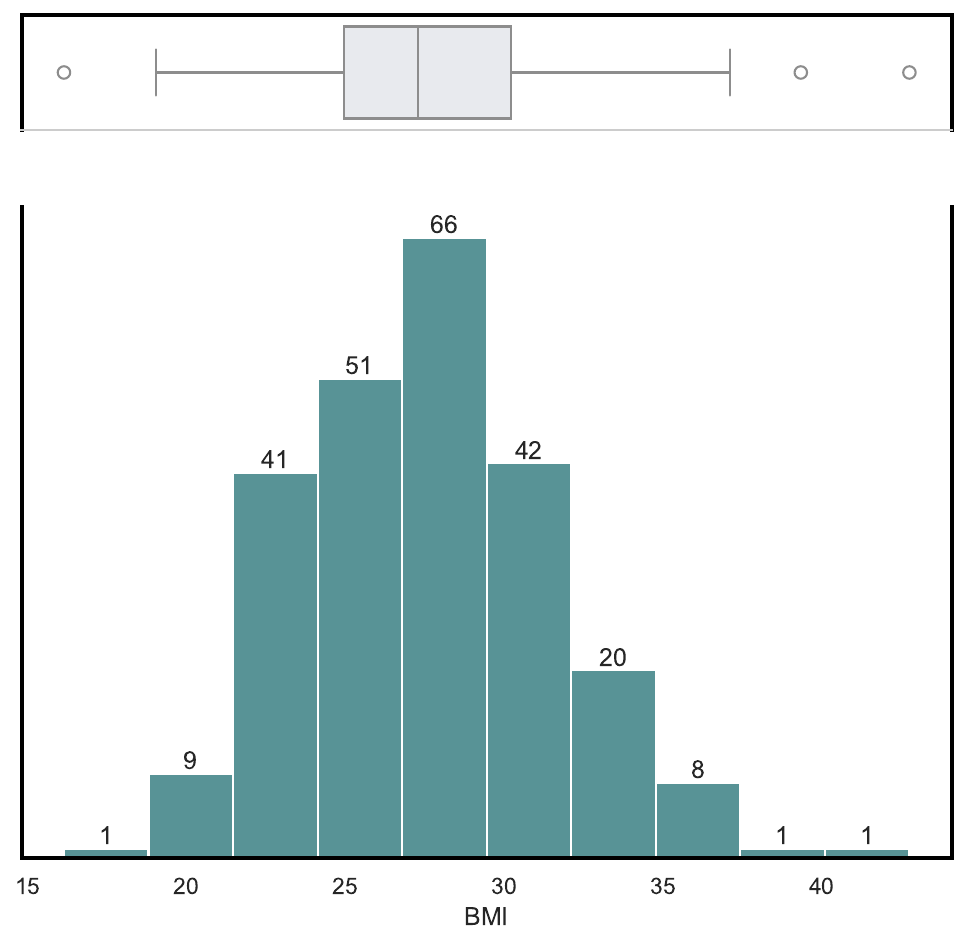}
       \subcaption{} 
    \end{subfigure}
    \begin{subfigure}{0.9 \textwidth}
        \includegraphics[width=0.98\linewidth]{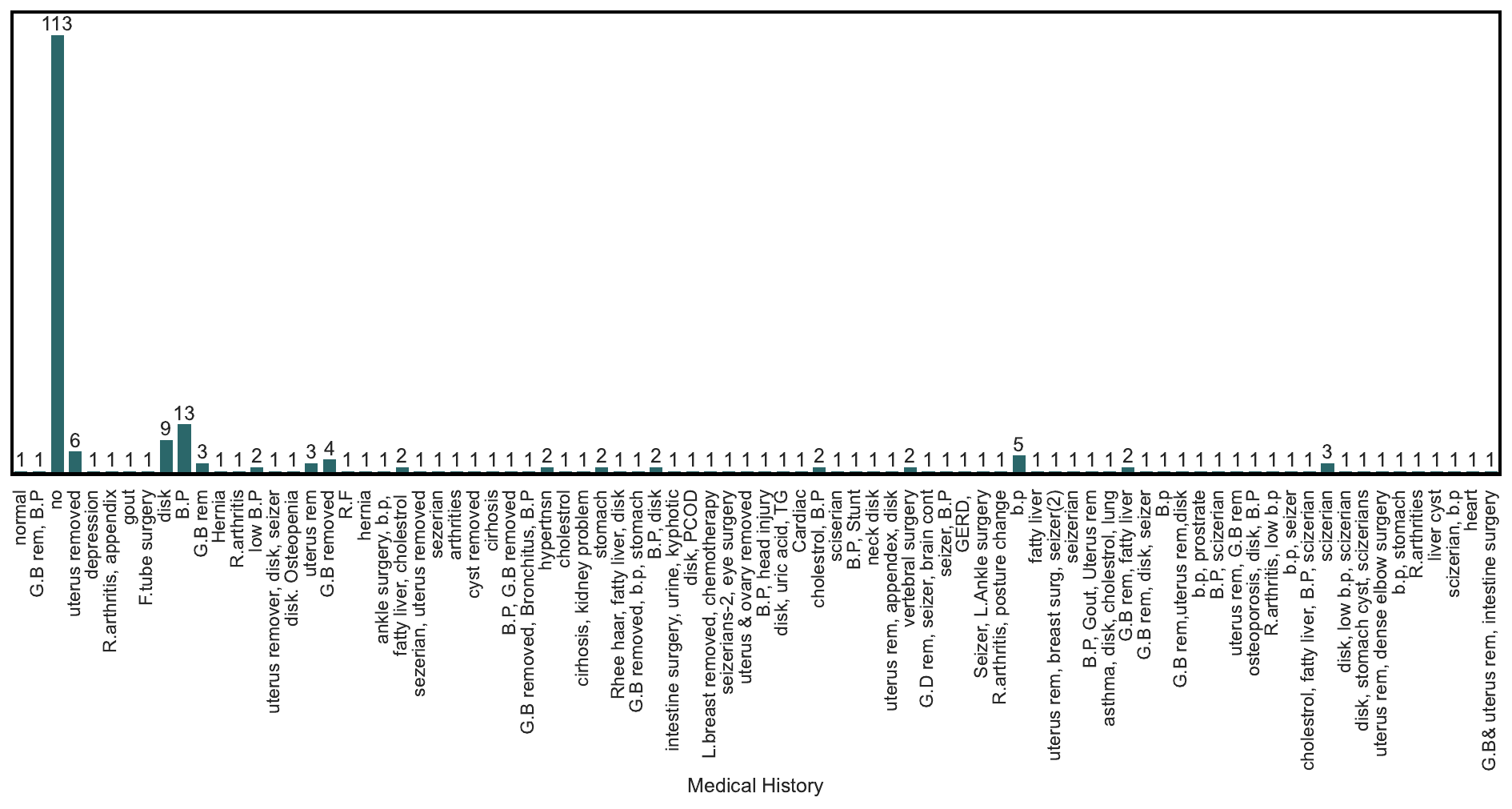}
       \subcaption{} 
    \end{subfigure}

    \caption{The statistical characteristics of the key continuous and categoric features of the screening data.}\label{screening data}
\end{figure*}

\section{Introduction}
\label{Introduction}
Osteoporosis is a common metabolic bone disorder in the elderly, which is characterized by a low bone mineral density (BMD) \cite{rasool2024konet, wani2023osteoporosis}. It often leads to bone pain and an increased risk of fragility fractures, which reduces the quality of life \cite{qiu2024developing, xiao2022global, wu2023prediction}. This condition has become an increasingly serious public health concern, particularly in countries with an aging population. Osteoporosis-related fractures, especially hip fractures, are among the leading causes of disability and mortality worldwide, imposing a significant social and economic burden on society \cite{qiu2024developing, zhu2022pulsed}. According to a systematic review and meta-analysis conducted in 2022, the global prevalence of osteoporosis and osteopenia was 19.7\% and 40.4\%, respectively \cite{xiao2022global}. Early identification of osteoporosis risk factors and timely referral to specialized care improve treatment outcomes and prognosis. Improving the diagnosis quality is crucial, because it promotes equity within the healthcare system by reducing waiting times for specialized care and it enhances the overall treatment process. Early diagnosis of osteoporosis enables the prevention of osteoporotic fractures and lowers costs to the public health system, since effective early-stage treatments can reduce expenses associated with surgeries and hospitalizations before fractures occur \cite{albuquerque2023osteoporosis, sarmadi2024comparative}. Although currently, dual-energy X-ray absorptiometry (DXA) is considered the gold standard for diagnosing osteoporosis, mass screening of individuals at high risk for osteoporosis limited. This limitation stems primarily from the high cost of DXA and its restricted availability in specialized hospitals, which makes the situation more severe in developing countries \cite{albuquerque2023osteoporosis}.

Replacing DXA with standard X-rays for osteoporosis diagnosis offers a more cost-effective and accessible alternative \cite{harvey2010osteoporosis, compston2010osteoporosis}. When compared with standard X-ray, DXA is superior because it directly measures BMD using two different energy levels of X-rays, allowing for precise detection of early-stage osteoporosis and small changes in bone mass. The technical complexity and the level of expertise necessary to formulate a diagnosis based on the imaging results contributes to the higher cost and limited availability of DXA. In contrast, standard X-rays are widely available and more affordable, but this technology lacks the sensitivity to detect early bone loss, relying instead on visual assessments of bone structure. To bridge this performance gap, advanced image processing techniques are necessary. To address this, researchers propose deep learning models to automatically extract relevant image features \cite{wani2023osteoporosis, rangayyan2024detection, gatineau2024development, dzierzak2022application}. By using these AI techniques, it is possible to enhance the diagnostic capability of standard X-rays, extracting detailed bone density information and making them a viable alternative to DXA \cite{wani2023osteoporosis, he2024deep}. Furthermore, this AI-driven approach can democratize osteoporosis screening and significantly reduce costs without sacrificing diagnostic accuracy \cite{luo2024deep}. However, decision support, delivered by deep learning models, is intransparent, because the decision system was created by tuning a large number of parameters during learning. In other words, you can recreate and understand the design steps of deep AI models, but retracing and validating the effects of learning on these models remains a challenge due to their complexity. This is a problem, because it is difficult to estimate or predict the effects of bias in the training data and in the selected deep learning algorithms. This makes it difficult to trust that the osteoporosis detection models can perform well in clinical settings.

While deep learning models have achieved impressive results in medical diagnostics \cite{he2024deep, chagahi2024cardiovascular}, their adoption in clinical settings faces challenges due to the black-box nature of these models. Explainable AI (XAI) approaches provide a solution by making model predictions more transparent and interpretable. In healthcare, where decisions must be understood and trusted by both professionals and patients, explainability is critical. Our approach focuses on enhancing explainability through feature selection and clustering, ensuring that every feature contributing to the final diagnosis is interpretable, thus promoting trust and transparency in the use of AI for osteoporosis detection. Our main contributions to this study are outlined below:

\begin{itemize}
    \item The data preprocessing involved cropping the primary tissues from the X-ray images and accurately removing irrelevant features from the screening data.

    \item  An ensemble transfer learning system is introduced that leverages three pre-trained networks, selected through their performances, to extract various features and patterns present in X-ray images.

    \item  By introducing a feature fusion layer, we combine features from pre-trained networks, eliminating null spaces and creating a more interpretable feature space. This enhances both performance and explainability, ensuring the model's decisions are more transparent and understandable.

    \item A component selection and synergistic combination (CSSC) layer carefully screen the components got from the feature fusion layer using a clustering-based variable selection approach. It combines these components with preprocessed clinical data, integrating hard and soft information to enhance transparency. 

    \item Cropping out X-ray images, utilizing pre-trained networks, employing feature fusion layers, and excluding redundant components contribute significantly to reducing computational costs. The synergistic feature combination achieves the desired performance metrics in osteoporosis detection.
\end{itemize}

The rest of this study is organized: Section \ref{Methodology} begins with an overview of the dataset and the identification of its key features, followed by a discussion on the feature fusion and component selection processes, and concludes with an analysis of the proposed system’s architecture. The findings and their analysis are detailed in Section \ref{Results}. Section \ref{recomm} discusses potential directions for future research. Last, Section \ref{Conclusion} provides the concluding remarks. 

\section{Methodology}
\label{Methodology}
In this study, we present a model that integrates Feature Fusion and Variable Clustering to enhance transparency and ensure effective, interpretable feature extraction from both X-ray images and tabular data. The model introduces two essential layers into the fully convolutional network classifier: (a) a fusion layer and (b) a Component Selection and Synergistic Combination (CSSC) layer. In the fusion layer, principal component analysis (PCA) is employed to reduce dimensionality while preserving the most relevant features. The CSSC layer then utilizes a clustering-based variable selection technique to identify and eliminate redundant features, combining the principal components with selected features from the screening data. This method not only optimizes model performance but also significantly improves interpretability, empowering healthcare professionals to better understand the factors influencing the model's predictions.

\begin{figure}[t]
\centering
\includegraphics[scale=0.5]{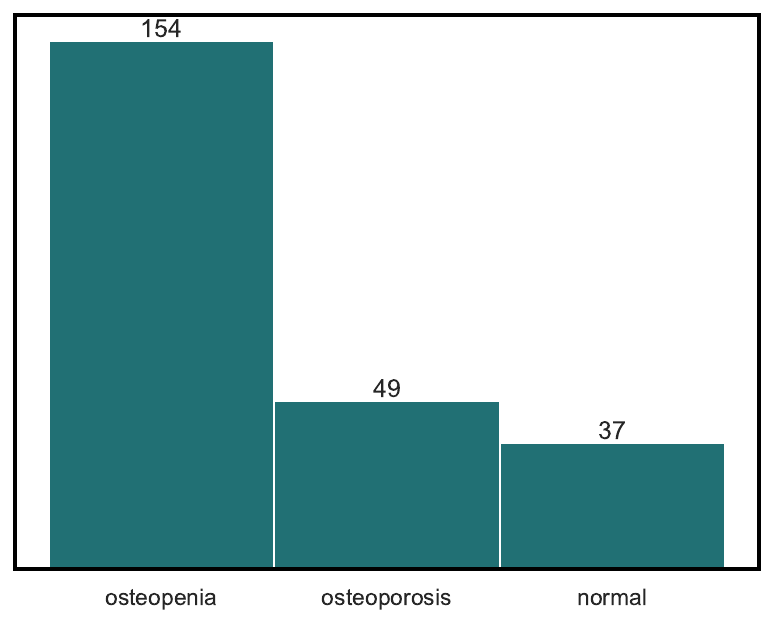}
\caption{Bar chart indicating of the number of samples in each class, highlighting the class imbalance.}
\label{target}
\end{figure}

\begin{figure*}[t!]
    \centering
    \begin{subfigure}{0.27 \textwidth}
        \includegraphics[width=\linewidth]{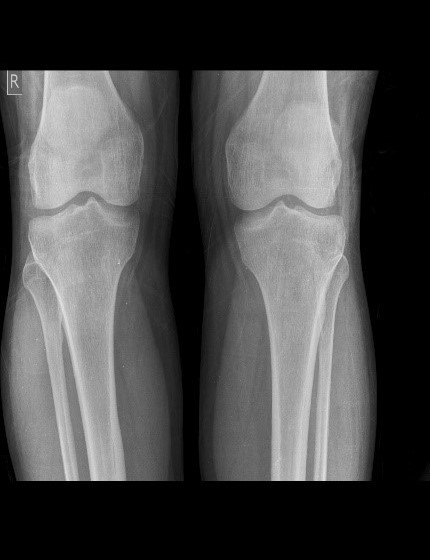}
        \subcaption{} 
    \end{subfigure}
    \begin{subfigure}{0.27 \textwidth}
        \includegraphics[width=\linewidth]{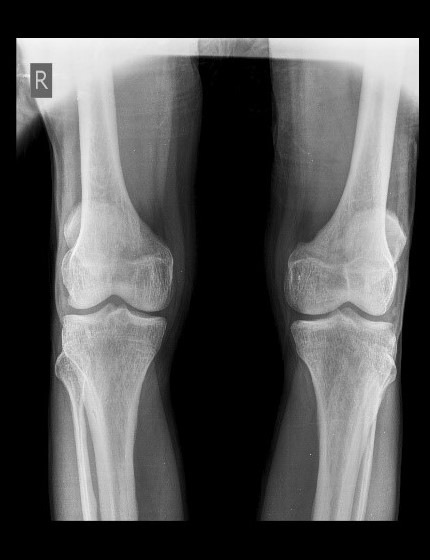}
        \subcaption{} 
    \end{subfigure}
    \begin{subfigure}{0.27 \textwidth}
       \includegraphics[width=\linewidth]{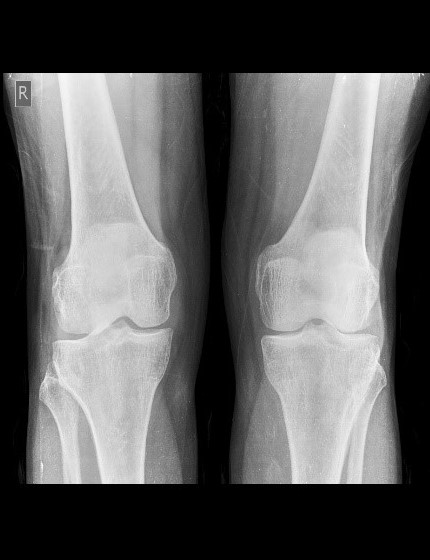}
       \subcaption{} 
    \end{subfigure}
    \caption{X-ray images illustrating the three bone density categories: a) normal, b) osteopenia, and c) osteoporosis. These images were taken from the labelled data used to train and test the osteoporosis detection models.}\label{x-ray}
\end{figure*}

\begin{figure*}[t]
    \centerline{\includegraphics[scale=0.4]{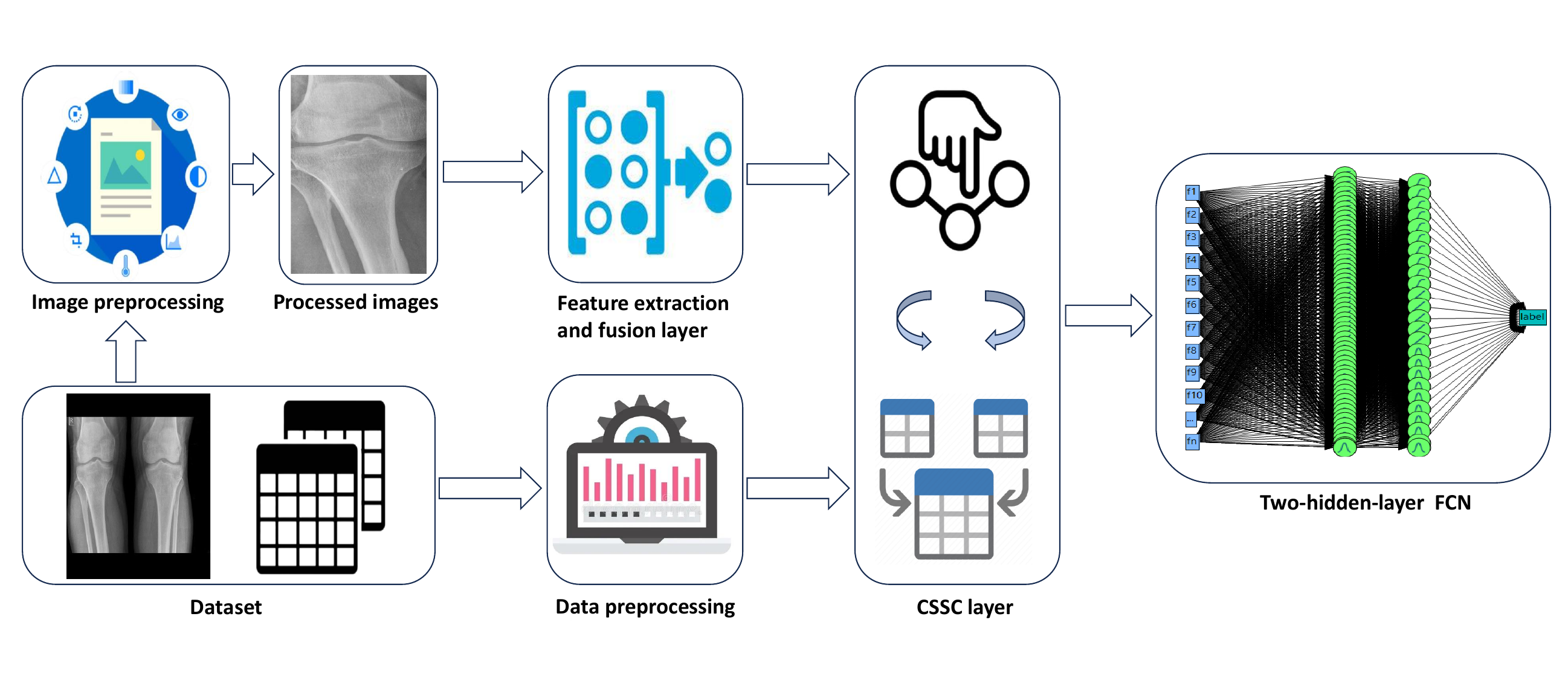}}
	\caption{Graphical overview of the proposed methodology for osteoporosis detection.}
	\label{graphical method}
\end{figure*}

\subsection{Dataset and Preprocessing}
\label{dataset}
The dataset used in this research was collected by the Unani and Panchakarma Hospital, Srinagar, J\&K, India \cite{kaggggle}.  It included X-ray images of the knee along with patient clinical data, covering variables such as gender, height (in meters), age, history of fractures, medical history, dialysis, joint pain, alcohol consumption, and site, which displayed identical values across different classes. Additional features, such as obesity, smoking habits, daily eating habits, diabetes, seizure disorders, family history of osteoporosis, and hypothyroidism, were also collected. However, these features were excluded as they were found to be uninformative based on feature importance analysis. Furthermore, variables with sparse data, such as the number of pregnancies and menopause age, were removed due to their limited availability across patients. Statistical details of the remaining key features, which were found to be most informative, are presented in Fig. \ref{screening data}. This study addresses a three-class classification problem, with the target categories being normal, osteopenia, and osteoporosis (\ref{target}).  Normal bone density reflects healthy bones with typical mineral levels. Osteopenia describes a condition where bone mineral density (BMD) is lower than normal but not yet low enough to be classified as osteoporosis. Osteoporosis represents a more severe reduction in BMD, resulting in brittle bones that are highly susceptible to fractures. Representative examples of X-ray images for each class are presented in Fig. \ref{x-ray}. To streamline processing and ensure the model focused on the most relevant areas, regions of interest (ROI) were identified in the X-ray images. For patients with images of both knees, we randomly selected the left or right knee for analysis. The preprocessed X-ray images were then fed into pre-trained models for feature extraction.

\begin{table}[t]
    \centering
    \caption{Single model performance on test data.}
    \begin{center}
    \begin{tabularx}{1.0\linewidth}{|l|>{\centering\arraybackslash}X|>{\centering\arraybackslash}X|>{\centering\arraybackslash}X|>{\centering\arraybackslash}X|>{\centering\arraybackslash}X|}
        \hline
        Model & Acc & Sens & Prec & Spec & F1 \\ \hline
        VGG19                       & 0.918 & 0.89   & 0.896 & 0.944 & 0.888 \\ 
        ResNet101v2-Inceptionv4      & 0.771 & 0.567  & 0.548 & 0.816 & NAN   \\ 
        ResNet50                     & 0.898 & 0.863  & 0.879 & 0.936 & 0.869 \\ 
        EfficientNet                 & 0.854 & 0.81   & 0.802 & 0.904 & 0.804 \\ 
        MobileNetV2                  & 0.83  & 0.698  & 0.827 & 0.87  & 0.742 \\ 
        InceptionV3                  & 0.894 & 0.831  & 0.874 & 0.92  & 0.848 \\ \hline
    \end{tabularx}%
    \end{center}
    \label{tab:model_performance}
\end{table}

\subsection{The Proposed Framework}
\label{proposed system}
The architecture developed for osteoporosis detection is illustrated in Fig. \ref{graphical method}. In this framework, preprocessed X-ray images are input into pre-trained models—VGG19, InceptionV3, and ResNet50—for feature extraction. These three networks were selected based on their individual performance, as evaluated on the test data and detailed in Table \ref{tab:model_performance}. The integration of these classifiers, coupled with a multi-modal learning approach, has led to substantial improvements in diagnostic accuracy and precision.diagnosis.

\subsubsection{Feature fusion layer}

Principal Component Analysis (PCA) is a widely used technique for dimensionality reduction that transforms high-dimensional data into a lower-dimensional space while preserving as much variance as possible.

In this study, we applied PCA to the features extracted from pre-trained networks, addressing the issue that these features may not be informative due to differences between the nature of our dataset and the datasets on which the models were trained. By performing PCA on the outputs of each pre-trained network, we retained the most important information from the images while reducing dimensionality. This approach ensures that the model concentrates on the most relevant features, preventing bias from non-informative dimensions. Subsequently, we concatenated the outputs of each low-dimensional feature set, thereby enhancing the model's capability with diverse feature representations.

\subsubsection{Component selection and synergistic combination layer}
Determining the optimal number of components to retain is a significant challenge in this research area. In this study, we adopted a criterion of selecting components that account for at least 2\% of the total variance of the extracted features to minimize information loss. Accordingly, we identified seven components for VGG19, seven for InceptionV3, and ten for ResNet50, all of which satisfied this requirement. While concatenating the principal components from these three models mitigates the risk of overlooking patterns in the X-ray images—since any pattern missed by one network may be captured by the others—it also introduces the potential for overfitting if identical patterns are identified across all networks. To address this concern and ensure diverse pattern capture while eliminating redundancy, we employed a clustering-based component selection approach. This approach clusters similar features based on correlation criteria, allowing us to select representative features from each cluster. By reducing redundancy and ensuring that the selected features convey diverse and non-overlapping information, we can enhance model performance. Table \ref{Summary} presents a comprehensive summary of the cluster characteristics, including the number of components within each cluster, the specific components of each cluster, the most representative component, and the proportion of variance explained both at the cluster level and overall. This table illustrates the effectiveness of the clustering approach in capturing variability within the dataset.

From the 24 components derived from the features extracted by the three pre-trained models, the variable clustering method selected one component from VGG19, three components from InceptionV3, and six components from ResNetV2-InceptionV3 as the most representative variables.

These selected variables were concatenated with the most informative features from the screening data to create a synergistic feature set. This new feature space, which includes the most informative features from the screening data alongside the carefully screened components from the X-ray images, serves as input to a fully connected network. Details of this network are provided in Table \ref{Model}.
\begin{table*}[!t]
    \centering
    \renewcommand{\arraystretch}{1.5}
    \caption{Summary of cluster characteristics and variance explained.}
    \label{Summary}
    \begin{tabular}{|c|c|c|c|c|c|}
        \hline
        Cluster & Number of & Cluster Variables & Most Representative & Cluster Proportion & Total Proportion \\
        & Members & & Variable & of Variation Explained & of Variation Explained \\
        \hline
        1 & 3 & PC1\_Inception, PC1\_VGG, PC2\_ResNet & PC1\_Inception & 0.688 & 0.086 \\
        \hline
        4 & 3 & PC3\_ResNet, PC2\_VGG, PC3\_Inception & PC3\_ResNet & 0.631 & 0.079 \\
        \hline
        2 & 3 & PC1\_ResNet, PC5\_Inception, PC5\_VGG & PC1\_ResNet & 0.597 & 0.075 \\
        \hline
        6 & 3 & PC3\_VGG, PC5\_ResNet, PC6\_Inception & PC3\_VGG & 0.575 & 0.072 \\
        \hline
        3 & 2 & PC6\_ResNet, PC4\_VGG & PC6\_ResNet & 0.788 & 0.066 \\
        \hline
        9 & 2 & PC9\_ResNet, PC6\_VGG & PC9\_ResNet & 0.753 & 0.063 \\
        \hline
        5 & 2 & PC2\_Inception, PC4\_ResNet & PC2\_Inception & 0.721 & 0.060 \\
        \hline
        8 & 3 & PC7\_Inception, PC7\_ResNet, PC7\_VGG & PC7\_Inception & 0.418 & 0.052 \\
        \hline
        7 & 2 & PC10\_ResNet, PC4\_Inception & PC10\_ResNet & 0.591 & 0.049 \\
        \hline
        10 & 1 & PC8\_ResNet & PC8\_ResNet & 1.000 & 0.042 \\
        \hline
    \end{tabular}\label{results1}
\end{table*}

\begin{table}[!t]
    \centering
    \renewcommand{\arraystretch}{1.3}
    \caption{shows the hyperparameters of a two-hidden-layer network.}
    \label{Model}
    \begin{tabular}{|c|c|c|c|c|c|c|}
        \hline
        \multirow{2}{*}{Hidden} & \multicolumn{3}{c|}{Activation Function} & \multirow{2}{*}{Learning} & \multirow{2}{*}{Epochs} & \multirow{2}{*}{Penalty} \\ 
        \cline{2-4} 
        Layers & Sigmoid & Identity & Radial & Rate &  & Method \\
        \hline
        $L_1$ & 25 & 10 & 25 & \multirow{2}{*}{0.1} & \multirow{2}{*}{100} & \multirow{2}{*}{Squared} \\
        $L_2$ & 10 & 5 & 10 &  &  & \\
        \hline
    \end{tabular}
\end{table}

\subsection{ Performance metrics}
Accuracy is the ratio of correctly predicted instances (both true positives and true negatives) to the total instances, e.g., 
\begin{equation}
   Accuracy = \frac{TP + TN}{TP + TN + FP + FN}
\end{equation}
where $TP$ is true positive, $TN$ is true negative, $FP$ is false positive, and $FN$ is false negative.

Sensitivity (also known as recall or true positive rate) is the ratio of correctly predicted positive instances to all actual positive instances, e.g., 
\begin{equation}
    Sensitivity = \frac{TP}{TP + FN}
\end{equation}

Precision (also known as positive predictive value) is the ratio of correctly predicted positive instances to the total predicted positive instances, e.g.,
   \begin{equation}
       Precision = \frac{TP}{TP + FP}
   \end{equation}

Specificity (also known as true negative rate) is the ratio of correctly predicted negative instances to all actual negative instances, e.g., 
   \begin{equation}
       Specificity = \frac{TN}{TN + FP}
   \end{equation}

F1-Score is the harmonic mean of precision and recall, providing a balance between the two metrics, useful for imbalanced datasets, e.g., 
   \begin{equation}
       F1-Score = \frac{2\times Precision \times Sensitivity}{Precision + Sensitivity}.
   \end{equation}
   
The Receiver Operating Characteristic (ROC) curve is a graphical representation used to evaluate the performance of a classification model. It illustrates the trade-off between the True Positive Rate (Sensitivity) and the False Positive Rate (1 - Specificity) at various threshold settings. AUC stands for “Area Under the ROC Curve.” It provides a single value that summarizes the performance of a classifier across all threshold values. The value of AUC ranges from 0 to 1.
\begin{itemize}
    \item $AUC = 1$: Perfect model; it perfectly separates the classes.
    \item $AUC = 0.5$: No discriminative power; equivalent to random guessing.
    \item $AUC < 0.5$: Indicates a model that performs worse than random guessing.
\end{itemize}

\section{Results}
\label{Results}
In this section, we will further examine the outcomes and provide various performance metrics to assess the efficiency of the suggested algorithm. These metrics encompass factors like accuracy, sensitivity, precision, specificity, and the F1-Score.

\begin{table}[t]
\centering
\caption{Performance measures on test set.}
\begin{tabular}{|l|>{\centering\arraybackslash}p{3cm}|}
\hline
Measure & Test Set\\
\hline
Generalized R-Square      & 0.9729 \\

Entropy R-Square          & 0.9307 \\

RASE                      & 0.0761 \\

Mean Absolute Deviation    & 0.0575 \\

Log-Likelihood            & -2.85  \\
\hline
\end{tabular} \label{Performance measures}
\end{table}

\begin{figure}[!t]
\centering
\includegraphics[scale= 0.4]{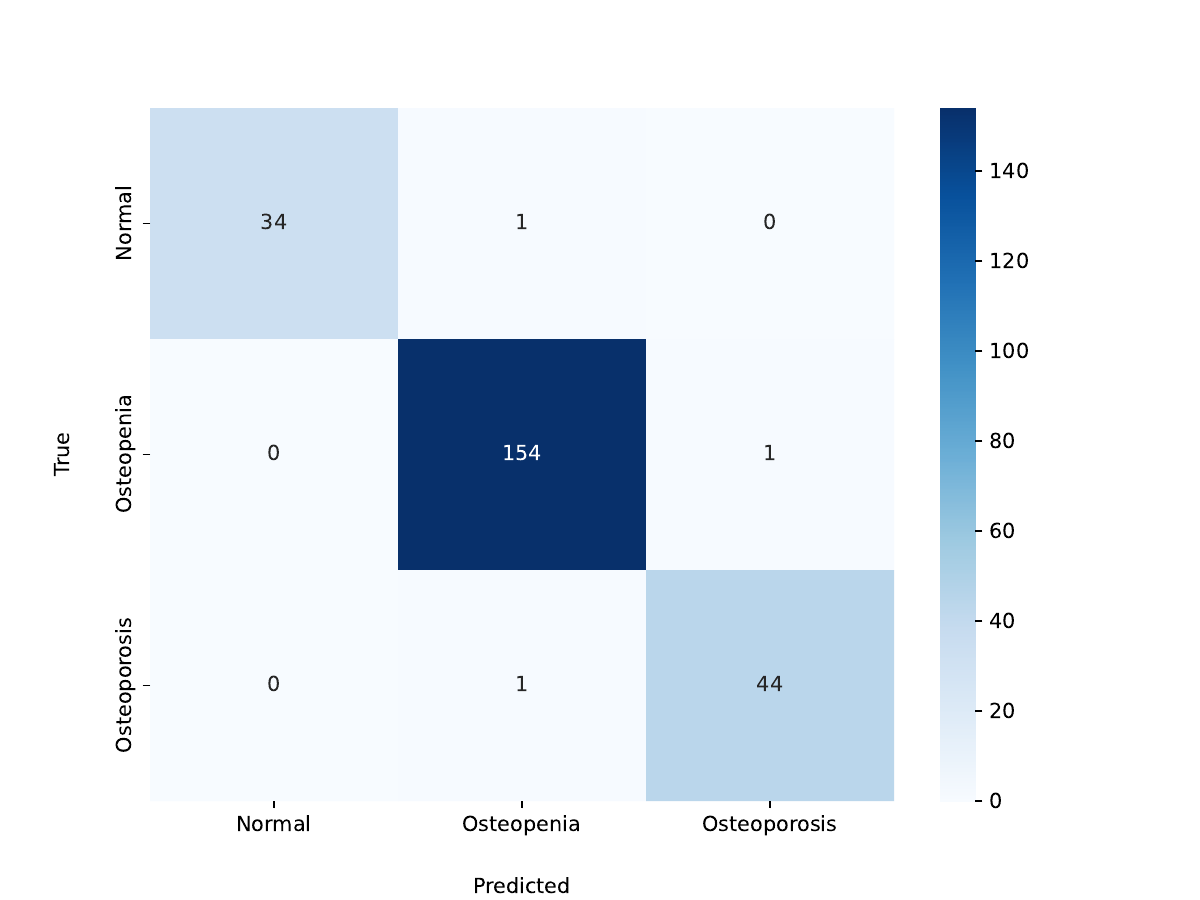}
\caption{Cumulative confusion matrix on test data.}
\label{test_confusion}
\end{figure}

\begin{figure}[!t]
\centering
\includegraphics[scale= 0.3]{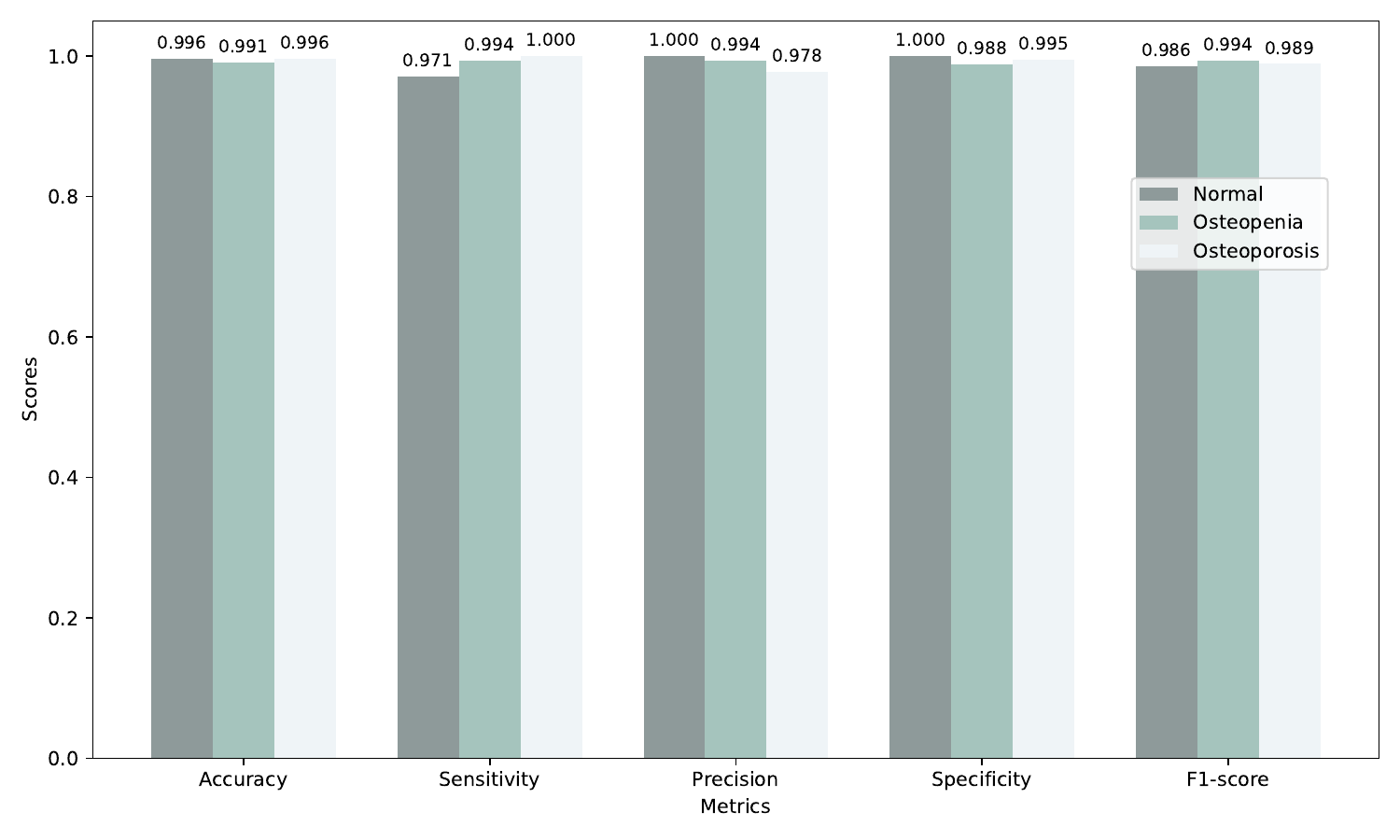}
\caption{Performance metrics on test data.}
\label{test_metrics}
\end{figure}

\begin{figure}[!t]
\centering
\includegraphics[scale= 0.45]{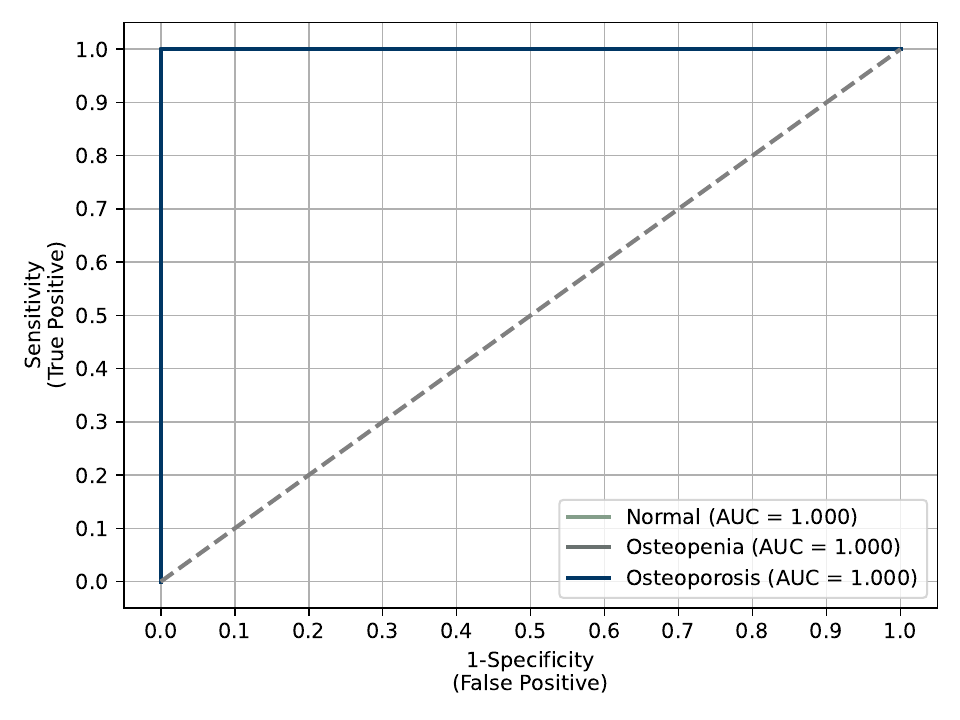}
\caption{ROC curve along with corresponding AUC on test data.}
\label{test roc}
\end{figure}

Table \ref{Performance measures} presents the performance metrics of the model evaluated on the test set. The Generalized R-Square value of 0.9729 highlights the model’s strong ability to explain variance and generalize to unseen data, indicating that the model captures the underlying patterns effectively. The Entropy R-Square, which reflects the model's confident predictions, also shows a high value of 0.9307, suggesting that the model is making accurate predictions with minimal uncertainty. The RASE (Root Average Squared Error) is 0.0761, demonstrating a low average squared error on the test set, which emphasizes the model’s precision in predicting outcomes. Furthermore, the Mean Absolute Deviation (MAD), calculated at 0.0575, shows that the average deviation between predicted and actual values is minimal, confirming the model’s robustness. Finally, the Log-Likelihood value of -2.85 suggests that the model is well-calibrated and fits the data well, reinforcing its reliability in providing accurate predictions for unseen data.\\
The performance of the proposed ensemble model on the test set is illustrated in Fig. \ref{test_confusion} and Fig. \ref{test_metrics}, with the results presented for the cumulative confusion matrix across the 5-fold cross-validation. Moreover, the ROC curve on test data, along with their corresponding AUC, is shown in Fig. \ref{test roc}.

\begin{figure}[!t]
\centering
\includegraphics[width=\linewidth]{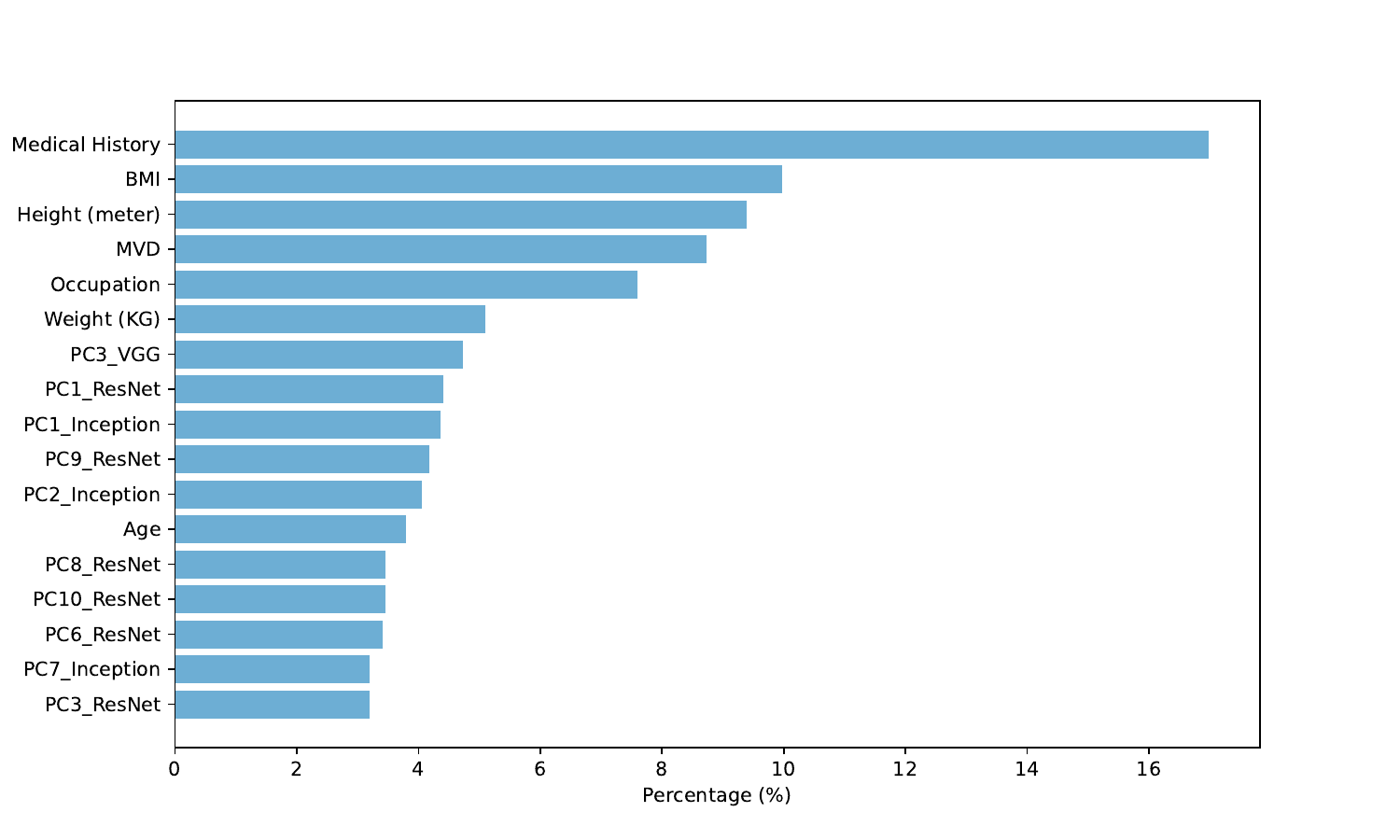}
\caption{Feature importance plot for the FCN model.}
\label{importance}
\end{figure}

Feature importance analysis was applied to the FCN model to assess the contribution of each input variable to the model's predictions. SHAP values were used to quantify feature importance, identifying the most important features for determining the model output. As shown in Fig. \ref{importance}, Medical History was the most significant feature, followed by BMI and Height (meter). These variables had the greatest impact on the ability of the model to classify the input. In contrast, features derived from pre-trained models, such as PC3-ResNet and PC7-Inception, showed a lower importance, indicating a smaller contribution to the final outcomes.

The feature importance plot highlights the relative influence of each input, with key features like Medical History and BMI playing a central role in determining predictions, while other features contributed minimally.

\section{Discussion}
Various methodologies and algorithms have been utilized to diagnose osteoporosis, which has seen a significant increase recently. These paragraphs explore several of these research endeavors.

Wani and Arora  \cite{wani2023osteoporosis} employed transfer learning such as AlexNet, VGG16, ResNet, and VGG19 to classify knee joint X-ray images into normal, osteopenia, and osteoporosis categories. The study compared classifier performance, highlighting that the pre-trained AlexNet architecture achieves the highest accuracy on their dataset, outperforming non-pretrained networks.

Rao et al. \cite{rao2022osteoporosis} employed five-level convolutional neural networks (CNNs) to detect osteoporosis in knee radiographs. They also incorporated clinical variables into ensemble models built on U-Net architectures. While the U-Net model demonstrated accurate osteoporosis identification from knee X-rays alone, incorporating patient data into the Efficient U-Net significantly improved performance metrics, such as accuracy, precision, and specificity. In contrast, GoogleNet, S-transform, and FCNN exhibited lower performance when applied to knee X-rays.

Kumar et al. \cite{kumar2023fuzzy} focus on classifying knee joint X-ray images into normal, osteopenia, and osteoporosis categories using three convolutional neural network architectures: InceptionV3, Xception, and ResNet 18. The study proposes an ensemble model that combines the outputs of these base classifiers through a fuzzy rank-based fusion technique, considering decision scores and confidence levels from each classifier. Unlike traditional fusion methods, this ensemble model bases its predictions on the confidence levels of the base learners, leading to more accurate results. The framework was evaluated using a 5-fold cross-validation method on a benchmark dataset.

Kumar et al. \cite{kumar2023enriched} concentrated on assessing the efficacy of various systems employed as feature extractors for differentiating between X-ray images of healthy individuals and those with osteoporosis. Their evaluation included testing knee images using eight renowned ImageNet pre-trained frameworks, specifically VGG16, VGG19, InceptionV3, Xception, DenseNet169, DenseNet201, ResNet101, and ResNet152.

Dzierżak and Omiotek \cite{dzierzak2022application} conducted a study that diagnosed osteoporosis using CT images of the spine. The images were classified into two categories: osteoporosis and normal. Six pre-trained DCNN architectures with varying topological depths, such as VGG16, VGG19, InceptionV3, Xception, ResNet50, and InceptionResNetV2, were employed in the analysis. Among these, the VGG16 model yielded the best results.

KONet, a robust deep learning model proposed by Rasool et al. \cite{rasool2024konet}, effectively differentiates between normal and osteoporotic knee conditions using medical images. To validate the ensemble approach, KONet was compared against several state-of-the-art CNN architectures: DenseNet-121, EfficientNetB0, ResNet50, VGG19, MobileNet, and InceptionV3. By leveraging transfer learning and combining the strengths of these models through a weighted ensemble, KONet significantly outperformed individual models, demonstrating its potential for accurate and reliable osteoporosis diagnosis.

Table \ref{Summary of Lit} presents a comparison between this study and other research utilizing deep neural networks for osteoporosis analysis. While previous studies have achieved commendable results, our model shows that by incorporating PCA and variable clustering, we can reduce the complexity of the model while improving its accuracy. More importantly, the model’s explainability is enhanced, as each selected feature can be linked to a specific medical insight, whether from X-ray images or patient screening data. This transparency is crucial for acquiring trust in AI-driven diagnostics, as it allows healthcare professionals to validate the reasoning behind each prediction.

\label{Summary of Lit}
    \setlength{\extrarowheight}{5pt}  
    \setlength{\arrayrulewidth}{1pt}  

\begin{table}[!t]
    \centering
    \renewcommand{\arraystretch}{1.5}
    \caption{Summary of comparison of the proposed method and other methods.}
    \label{Summary of Lit}
    \renewcommand{\arraystretch}{0.8}
    {\small
    \begin{adjustbox}{width=0.5\textwidth}
    \begin{tabular}{|c|c|c|c|c|c|c|c|}
        \hline
        Work & Dataset & Modality & Number of & Objective & Methods & Validation & Performance \\
        & & & Samples & & & Strategy & \\
        \hline
         &  &  & &  &  &  & Acc = 0.911 \\
        \cite{wani2023osteoporosis} & Private & knee X-ray & 381 & Multi-class & AlexNet & 1-fold & Error rate = 0.09 \\
         &  &  & &  &  &  & Val loss = 0.54 \\
        \hline
        &  &  &  &  &  &  &  Acc = 0.928\\
        &  &  &  &  &  &  &  Sens = 0.874\\
        \cite{pan2024effectiveness} & Private & chest CT & 1048 & Multi-class & ResNet-101 residual DCNN & 1-fold & Spec = 0.941  \\
         &  &  &  &  &  &  & AUC = 0.974\\
        \hline
         &  &  &  &  & Ensemble model &  & Acc = 0.935 \\
        \cite{kumar2023fuzzy} & Public & knee X-ray & 240 & Multi-class &( [Inception v3, Xception, & 5-fold & Val loss = 0.082 \\
         &  &  &  &  & Resnet] + fuzzy rank unification) &  &  \\
        \hline
         &  &  & &  &  &  & Acc = 0.8636 \\
        \cite{kumar2023enriched} &  Public & knee X-ray & 372 & Binary & VGG16 & 1-fold & Prec = 0.8696 \\
         &  &  &  &  &  &  & Sens = 0.8636 \\
          &  &  & &  &  &  & F1-Score = 0.8631 \\
        \hline
         &  & & &  & &  & Acc = 0.95 \\
        \cite{dzierzak2022application} & Private &  L1 spongy & 400 & Binary & VGG16 & 1-fold & TPR = 0.96 \\
         &  & tissue CT & & &  &  & TNR = 0.94 \\
        \hline
         &  &  & & & Ensemble model (Resnet50,  &  & Prec = 0.97 \\
        \cite{rasool2024konet} & Public & knee X-ray & 372 & Binary & VGG19, DenseNet121, MobileNet & 5-fold & Sens = 0.97 \\
         &  &  & & & EfficientNetB0, InceptionV3) &  & F1-Score = 0.97 \\
        \hline
        &  & CT-scans data &  & &  &  & Acc = 0.946 \\
        \cite{oh2024evaluation} & Private & from chest, spine, & 112 & Multi-class & DL-BMD & 1-fold & Sens = 0.955 \\
         &  & abdominal &  & &  &  & Spec = 0.935 \\
        \hline
        &  &  & & & &  & Acc = 0.74 \\
        \cite{feng2023deep} & Private & hip X-ray & 139 & Multi-class & DenseNet121  & 10-fold & Sens = 0.68\\
        &  &  & & & &  & Spec = 0.80 \\
        &  &  & & & &  & F1-score = 0.71 \\
        \hline
        &  &  & & & &  &  Acc = 0.933 \\
        \cite{wani2024deep} & Private & knee X-ray & 240 & Multi-class & Ensmble model & 1-fold & Error rate = 0.06 \\
         &  &  & & & (Alexnet, ResNet) &  & \\
        \hline
        &  & CT-scans &  & & &  & Acc = 0.909 \\
        &  & data from &  & & &  & Prec = 0.899 \\
        \cite{peng2024study} & Private & chest, & 1219 & Multi-class & DenseNet & 1-fold & Sens = 0.908 \\
        &  & abdomen, &  & & &  & Spec = 0.956 \\
        &  & and spine &  & &  &  & F1-score = 0.903 \\
        \hline
        &  &  &  &  & & & Acc = 0.638 \\
        \cite{sarmadi2024comparative} & Public & knee X-ray & 240 & Multi-class & Vision transformer & 5-fold & Prec = 0.6283 \\
        &  &  & & &  &  & Sens = 0.6383 \\
        &  &  & & &  &  & F1-score = 0.6067 \\
        \hline
        &  &  & & &  &  & Acc = 0.991\\
        &  &  &  & &  &  & Prec = 0.991 \\
        Our study & Public & knee X-ray & 240 & Multi-class & Deep ensemble model (VGG19, & 5-fold & Sens = 0.988 \\
        &  &  & & & Inception v3, Resnet50) &  & Spec = 0.994 \\
        &  &  &  & & &  & F1-score = 0.989 \\
        &  &  &  & & &  & AUC = 1.0 \\
        \hline
    \end{tabular}
    \end{adjustbox}
    }
\end{table}

\subsection{Future Research Directions}
\label{recomm}
In the future, we aim to explore:
\begin{itemize}
    \item In this study, the selection of classifiers for combination was based on their performance on the test data. In future research, the selection of classifiers can be based on factors such as the type of data, the diversity of extracted features, interpretability, and other criteria.
    
    \item Dimensionality reduction in this study was performed using PCA. Other fusion approaches, such as t-distributed stochastic neighbor embedding, linear discriminant analysis, and independent component analysis, can be considered in future research.
    \item Future research could explore more advanced explainability techniques to further enhance the interpretability and trustworthiness of AI in medical diagnostics.

    \item We selected components that accounted for over 2\% of the data variance as informative features. Future researchers can focus on optimizing the number of components by designing appropriate mechanisms.

    \item We performed a synergistic combination by concatenating features got from X-ray images with selected features from the screening data. Techniques such as feature union and feature stacking, among others, can be explored as alternatives to concatenation in the future.
\end{itemize}

\section{Conclusion}
\label{Conclusion}
Early diagnosis of osteoporosis is crucial for several reasons, including fracture prevention, reducing treatment costs, and avoiding the loss of mobility. This research, aware of the real challenges faced by healthcare centers such as the lack of labeled data and medical image processing problems, introduced a new synergistic-based network that effectively addresses these issues.

Our proposed model utilized three pre-trained models—VGG19, InceptionV3, and ResNet50—to extract diverse patterns from X-ray images. The use of PCA transformation from ${R}^n$ to ${R}^m$ where $n \gg m$ , helped eliminate null spaces and focus the model on the most informative features. The feature importance analysis revealed that Medical History, BMI, and Height (meter) were the most critical features, indicating that patient-specific medical and physical characteristics play a larger role in the predictions than image-derived components. The pre-trained model features, such as PC3-ResNet and PC7-Inception, contributed less to the final predictions, highlighting that while image-based features are useful, clinical data remains central to accurate diagnosis.

By combining the most representative components obtained through clustering-based selection with preprocessed screening data, the model fed an FCN for final labeling. This approach not only achieves high accuracy but also enhances explainability by emphasizing medically relevant factors in its predictions. This transparency is key for building trust among healthcare providers and patients, supporting the integration of AI models into clinical practice while ensuring that decisions are based on understandable and meaningful inputs.
\bibliography{references}

\end{document}